\def\year{2021}\relax
\documentclass[letterpaper]{article} 
\usepackage{aaai21}  
\usepackage{times}  
\usepackage{helvet} 
\usepackage{courier}  
\usepackage[hyphens]{url}  
\usepackage{graphicx} 
\def\UrlFont{\rm}  
\usepackage{natbib}  
\usepackage{caption} 
\usepackage{boldline,multirow}
\usepackage{ragged2e}
\usepackage{amsmath}
\usepackage{array}
\title{Data Representing Ground-Truth Explanations to Evaluate XAI Methods }
\author{
    Shideh Amiri\textsuperscript{\rm 1}, 
    Rosina O. Weber\textsuperscript{\rm 2}, 
    Prateek Goel\textsuperscript{\rm 2}, 
    Owen Brooks\textsuperscript{\rm 3}, 
    Archer Gandley\textsuperscript{\rm 3}, 
    Brian Kitchell\textsuperscript{\rm 3}, 
    Aaron Zehm\textsuperscript{\rm 3}\\
}
\affiliations {
    \textsuperscript{\rm 1}Department of Civil, Architectural and Environmental Engineering, Drexel University \\
    \textsuperscript{\rm 2}Information Science, Drexel University \\
      \textsuperscript{\rm 3}Computer Science, Drexel University \\
    {ss4588, rw37, pg427, oqb23, alg383, bgk32, amz52}@drexel.edu\\ 
      \textsuperscript{\rm 3}archergandley@gmail.com,  briankitchellxd@gmail.com \\
}

\begin{document}

\maketitle

\begin{abstract}
Explainable artificial intelligence (XAI) methods are currently evaluated with approaches mostly originated in interpretable machine learning (IML) research that focus on understanding models such as comparison against existing attribution approaches, sensitivity analyses, gold set of features, axioms, or through demonstration of images. There are problems with these methods such as that they do not indicate where current XAI approaches fail to guide investigations towards consistent progress of the field. They do not measure accuracy in support of accountable decisions, and it is practically impossible to determine whether one XAI method is better than the other or what the weaknesses of existing models are, leaving researchers without guidance on which research questions will advance the field. Other fields usually utilize ground-truth data and create benchmarks. Data representing ground-truth explanations is not typically used in XAI or IML. One reason is that explanations are subjective, in the sense that an explanation that satisfies one user may not satisfy another. To overcome these problems, we propose to represent explanations with canonical equations that can be used to evaluate the accuracy of XAI methods. The contributions of this paper include a methodology to create synthetic data representing ground-truth explanations, three data sets, an evaluation of LIME using these data sets, and a preliminary analysis of the challenges and potential benefits in using these data to evaluate existing XAI approaches. Evaluation methods based on human-centric studies are outside the scope of this paper. 
\end{abstract}


\section{Introduction}
Research in interpretable machine learning (IML) has explored means to understand how machine learning (ML) models work since the nineties ({\it e.g}., \citeauthor{towell1992interpretation} \citeyear{towell1992interpretation}). Popular methods to help understand ML models are referred to as attribution methods \citep{olah, yeh2018, koh2017understanding}, they identify features or instances responsible for a classification.

With the exception of human-centered studies \citep{hoffman2018metrics}, the evaluation methods being used in XAI and  IML include comparison to existing methods, metrics and axioms, sensitivity analyses, gold features, and demonstration of image classification (details and references in Section Background and Related Works). The problems with these methods include that they do not indicate where current XAI approaches fail thereby preventing consistent progress of the field. They do not measure accuracy as a way to validate correctness or to produce accountable agents ({\it e.g}., \citeauthor{Diakopoulos2014AlgorithmicAR} \citeyear{Diakopoulos2014AlgorithmicAR}, \citeauthor{kroll2016accountable} \citeyear{kroll2016accountable}, \citeauthor{doshivelez2017rigorous} \citeyear{doshivelez2017rigorous}), and it is practically impossible to determine whether one XAI method is better than other or what the weaknesses of existing methods are,  leaving  researchers  without  guidance  on  which  re-search questions will advance the field.

The intended purpose of this paper is to address these limitations with current XAI evaluation methods by proposing the use of data representing ground-truth explanations (GTE). In a variety of computer science tasks, it is a standard practice to treat some representation of data as ground truth. Ground-truth data, in its essence, represents data that is verifiable and considered as the most accurate against which a new system is tested against \citep{wikilink}. Various authors agree that the lack of ground-truth for evaluating explanations is a limitation \citep{tomsett2019sanity, hooker2019benchmark, yang2019evaluating, Montavon2019}. Consequently, we investigate the challenges in creating data representing GTEs. Our goal is to promote consistent and methodical progress of the XAI field. The scope of this paper is limited to neural networks (NN) for classification tasks.

The next section presents related methods, metrics and axioms to evaluate XAI methods. Then, we introduce how to generate three data sets representing GTEs. Next, we use the generated data to train NN models to submit to LIME \citep{ribeiro2016should} and produce explanations while converting the GTEs into LIME's explanations format. We evaluate LIME, and analyze the evaluation, seeking support to our conclusions as a means to validate the evaluation. We conclude with a discussion of issues and benefits, and future work.
\begin{table*}[t]
\centering
\small
\resizebox{\textwidth}{!}{
\begin{tabular}{p{0.31\textwidth}|p{0.69\textwidth}}
    \hline
    \textbf{Evaluation method/Axiom/Metric} & \textbf{Method proposer and/or example authors who employed them}\\
    \hline
    Sensitivity analysis &  \citet{adebayo2018sanity}\\
    \hline
    Example images &  \citet{ribeiro2016should}\\
    \hline
    SAT/Model counting & \citet{ignatiev2018, narodytska2019assessing, ignatiev2019validating} \\
    \hline
    Correlation, completeness, and complexity &  \citet{cui2019integrative}\\
    \hline
    Conservation, continuity &  \citet{Montavon2019}\\
    \hline
    Fidelity & \citet{alavarez2018}\\
    \hline
    Gold features &  \citet{ribeiro2016should}\\
    \hline
    Post-hoc accuracy  & \citet{chen2018learning, Bhatt2019BuildingHT, xie2019reparameterizable, bai2020attention}\\
    \hline
    Perturbation analysis for vision   &  \citet{zeiler2014visualizing}\\
    \hline
    ROAR    &  \citet{hooker2019benchmark}\\
    \hline
    Perturbation on Time Series     &  \citet{schlegel2019rigorous}\\
    \hline
    Implementation invariance, sensitivity      &  \citet{sundararajan2017axiomatic}\\
    \hline
    Input invariance      &  \citet{kindermans2019reliability}\\
    \hline
    Simulated users            &  \citet{ribeiro2016should}\\
    \hline
    Amazon Mechanical Turk users                  &  \citet{ribeiro2016should, chen2018learning}\\
    \hline
    In-depth Interviews & \citet{hoffman2018metrics}\\
    \hline
\end{tabular}
}
\caption{Methods, metrics or axioms used to evaluate XAI and IML methods}
\label{alternativetable1}
\end{table*}
\section{Background and Related Works} \label{BGDRW}
Table 1 lists various methods currently used to evaluate both XAI and IML methods. The authors referenced in the table are those that first proposed the methods or that have used them within XAI and IML works. None of these methods use data representing ground-truth explanations. The closest to ground truth is the use of gold features \citep{ribeiro2016should}, which are a set of features used in a model that are well-known to be the most important.

\citet{doshivelez2017rigorous} categorize IML evaluations as application-, human-, and functionally-grounded. The authors propose that any method should be evaluated along those three categories, one informing the other. \citet{yang2019evaluating} are the only authors who actually present a reason against using ground truth to benchmark explanation methods, which is that of explanation quality is user-dependent. These authors propose three metrics for IML, namely, generalizability, fidelity, and persuasibility. Their fidelity metric aims to measure explanation relevance in the applied context. \citet{gunning2019darpa} propose XAI approaches are evaluated along five categories, namely, {\it explanation goodness}, {\it explanation satisfaction}, {\it mental model understanding}, {\it user task performance}, and {\it appropriate trust and reliance}. Considering that human-centered studies entail a lot of subjectivity, of those, only {\it explanation goodness} seems as an objective category of explanation quality. All other categories are evaluated by humans or an external task. \citet{tomsett2019sanity} conducted a meta evaluation of saliency methods by analyzing metrics previously proposed in the literature to evaluate saliency methods. To do this, they adopted psychometric tests to verify local saliency metric reliability and consistency of the saliency metrics that rely on perturbations. They conclude that saliency metrics can produce unreliable and inconsistent results, even if not always.

\begin{figure}[ht]
\centerline{\includegraphics[width=\columnwidth, height=4cm]{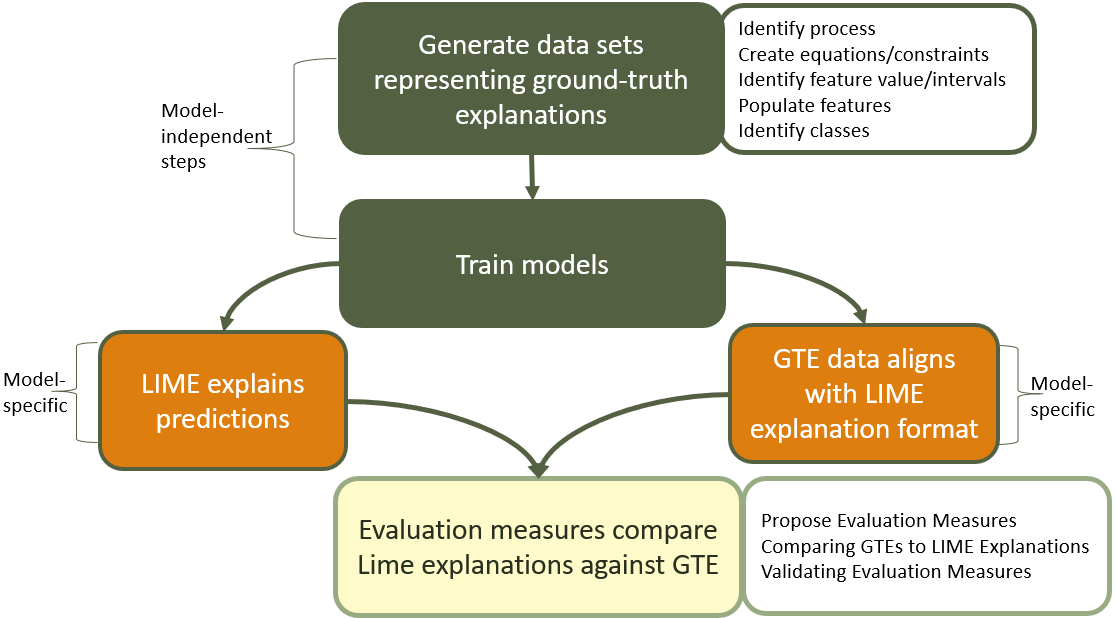}} 
\caption{{In green, this diagram shows the steps to generate data representing GTEs and to use the data to train models. In orange are the processes for aligning GTE to LIME, and to send predictions for LIME to explain. In yellow, the evaluation compares the two orange processes}}
\label{gted}
\end{figure}

\section{Generate Data Representing Ground-Truth Explanations (GTE)}
Figure \ref{gted} gives an overview of the entire approach from generating the data up to evaluation. We start describing how to generate the data. We propose to generate data sets from existing processes, either natural or artificial, and identify classes from said processes. We propose to represent classes in a data set via mathematical equality or inequality constraints as a minimal canonical representation from which explanations can be aligned with the format of explanations produced by various XAI methods. We define the classes and the intervals to populate feature values to create instances. The intervals where instance feature values can be populated will determine noise and commonsense. The nature of values allowed for each feature in the generated equations will determine whether classes will remain disjoint. Overlapping classes will impact evaluations producing noise. Another consideration when defining intervals to populate feature values is commonsense. If an explanation indicates that the value of a feature is 0.3x$10^{-6}$, then the feature should not represent someone's age. Next, we describe the generation of three data sets. 

\subsection{Generate Data Set Loan}

The process we chose is loan underwriting. This is a small data set, consisting of 54 instances, two classes accept and reject, and three input features characterizing three well-known factors considered in loan underwriting, namely, job condition, credit score, and debt-to-income ratio. We created this data set manually to characterize realistic instances. 
The instance space is given by the arrangement of the three features and their four possible values, given by 4 x 4 x 4 = 64. We eliminated 10 instances from the data because they were not realistic. The data is generated with a system of two equations as follows:
\begin{equation}
    f(x)= 
\begin{cases}
    8(x_1 - 2)^2 + 3x_2^3 - x_3^4 + 4, & \text{\it if } x\ne 2\\
     3x_2^3 + x_3^4 + 12,              & \text{\it if } x= 2
\end{cases}
\end{equation}
\begin{equation}
\begin{cases}
    Accepted, & \text{\it if } f(x)\ge 32\\
    Rejected,              & \text{\it if } f(x)< 32
\end{cases}
\end{equation}

As stated above, we considered class overlap and commonsense when defining the allowable values for the three features. The first feature, $x_1$ corresponds to the job condition of the applicant. This feature can be populated with integer values along the interval [2, 5], where 2 represents lack of a job, and values 3, 4, and 5, respectively that applicant has a job for less than one year, less than 3 years, or more than 3 years. The second feature, $x_2$, refers to credit score, which assumes integer values in the interval [0, 3], distributed in ranges from less than 580, 650, 750 and more than 750. The third and last feature, $x_3$, refers to the ratio of debt payments to monthly income, which assumes integer values in the interval [0, 3], distributed in ranges from less than 25\%, 50\%, 75\% and more than 75\%.

\subsection{Generate Data Set Distance}

We adopt the equation used to calculate travel consumption based on travel distance. The Data Set Distance has a total of 2,600,000 instances, described through five features, and 10 classes.
The 5 variables are Trip Frequency (TF), Trip Distance (TD), Transportation Occupancy (TO) ,Energy Intensity (EI) and Travel Mode (m). The Data Set Distance is generated using Equation 4 for travel energy consumption based on travel distance. 

Using the base equation, we created 10 unique variations with the following goals: the variations should be kept realistic, the variations are a set of operations (such as raising to an exponent or multiplying by a scalar) performed on one or more variables. Afterward, we  generate the data for the base equation by creating every permutation of 4 equation variables within a specified range using a truncated normal distribution. The 4 variables are used as features along with a 5th variable, travel mode. For each of the 10 variations, we use the set of operations on the base equation data to generate the equivalent rows for that variation.
\begin{equation}
    E_m = TF \times \frac{TD_m}{TO_m} \times EI_m
\end{equation}

\subsection{Generate Data Set Time}

For Data Set Time and Distance, we used processes from the field of energy consumption analysis that describe various realistic processes with different focuses (e.g., distance or time) and include equations with a variety of features that can receive multiple values. These characteristics facilitate the generation of large data sets, so we can create conditions similar to those faced by XAI methods in the real world. In this paper, we generate data from transportation energy consumption, which can be used to calculate travel time and travel distances related to household energy consumption.

Equation 3 is the basic equation to calculate energy (E) as a function of time. The four variables are Travel Time (TT), Speed, Fuel Economy (FE) and Travel Mode (m). The Data Set Time has a total of 504,000 instances and seven classes. Each class is defined by a small tweak to the equation. Using the base equation, we create seven unique variations with the same goals and process as we did for the Distance data set.

\begin{equation}
    E_m  = \sum_{m=1}^{5} = TT \times Speed_m \times FE_m
\end{equation}

\subsection{Train NN Models}
The number of models, the type of models, and how they vary between them depends on the metrics selected in the previous step. Consider, for example, the selected metric is {\it implementation invariance} \cite{sundararajan2017axiomatic}. This metric requires multiple types of models. In this paper, we trained two models for the Loan and Time data sets, and one model for the Distance data set, which we summarize next (detailed architectures are given in the Github link given at the end of this paper).

\subsubsection{Models NN1 and NN2}
The changes from NN1 to NN2 for Loan and Time included number and type of layers. Both models built for the Loan data reached 100\% accuracy. Given the small number of 54 instances, we did not separate testing and training. For Data Set Time, NN1 reached 97\% accuracy and NN2 96\%. The generated instances were set to 403,200 for training, and 100,800 for testing.
Train NN 2 to 96\% accuracy 403,200 training instances, 100,800 testing instances. The accuracy obtained for Data Set Distance was much lower, 82\%. This is certainly due to noise class overlaps that occurred during the data set generation. 

\subsection{LIME Explains Predictions}
As depicted in the diagram in Figure \ref{gted}, after training the models, the next steps can be concurrent. This section describes the step where LIME explains the predictions from the models. First, let us briefly review how LIME works. The Local interpretable model-agnostic explanations (LIME) is a feature attribution method formally described in \citet{ribeiro2016should}. LIME assumes that the behavior of a sample point ({\it e.g}., instance) can be explained by fitting a linear regression with the point (to be explained) and the point's close neighbors. LIME perturbs the values of the point to be explained and submits them to the model to obtain their prediction, thus creating its own data set. Next, LIME measures the cosine similarity between the point to be explained and the points generated from its perturbations to select a region of points around it. LIME then utilizes a hyperparameter, {\it number of samples (num\_sample)}, to select the number of points it will use in the final step, which is the fitting of a linear regression. The hyperparameter {\it num\_sample} determines how many of the perturbed points will be used with the point to be explained to fit a linear regression. This last step produces coefficients of the line that expresses LIME's explanation. 

For Data Set Loan, we submit to LIME all the 54 instances to be explained, and models NN1 and NN2. Note that all these instances with both NN models will have correct predictions because both models reached 100\% accuracy. The number of samples selected was 25 to be used in the first evaluations, but we also created GTE for 5 and 50 number of samples. The output we receive back from LIME are two sets of 54 by 3 coefficients, one coefficient for each of the three features, one set for NN1, and one set for NN2. 

The Data Set Time has 100,800 instances but the models did not reach 100\% accuracy. Consequently, we randomly selected 10,000 that both models NN1 and NN2 predicted correctly and only submitted those 10,000 to LIME together with the two models. We made sure to select the instances from those correctly predicted because sending instances incorrectly predicted by the models would mislead LIME to produce bad decisions, and we had to make sure we could provide LIME with the best data for a fair assessment. The output produced by LIME are two sets of 10,000 by 4 coefficients, accounting for the four features in this data set, one for each model NN1 and NN2. We set hyperparameter number of samples as 1,000.

The accuracy reached by the Distance model NN1 was 82\%. The number of testing instances was 520,000, so we randomly selected 50,000 instances from this data set from those correctly predicted. The data produced by LIME for NN1 is a 50,000 by 5 matrix, given the five features in this set. For number of samples, we used 5,000.

\subsection{GTE Data Aligns with LIME Explanation Format}

This step is represented in Figure \ref{gted}, in orange, and is concurrent to the step "Lime Explains Predictions". As already noted, the data we produce representing GTEs are a specific way to represent explanations. We can only use them to evaluate any target XAI method after the data representing GTEs are at the same format or have been processed under the same conditions. As a general rule of thumb, this conversion may imply to take the ground-truth data and execute the last steps of the target method.

For LIME, an explanation consists of a fitted line from a target point whose prediction we want to explain and the number of points (based on the hyperparameter number of samples, {\it num\_sample}) that are the closest to the target based on the cosine similarity. Ultimately, this means that evaluating LIME means to determine how realistic are the perturbed points that are the closest based on cosine similarity in the number of the number of samples. Consequently, we take the points from the data representing GTEs and execute this same process, namely, measure the cosine similarity of each target point to be explained and then fit linear functions using the same Ridge regression method and the same regression parameters used in LIME using the same number of samples as established by the {\it num\_sample} hyperparameter. The result is that, for each data set, we have matrices with the same number of coefficients (one per feature per instance) as produced by each model. 

\section{Evaluation Measures Compare Lime Explanations Against GTE}
\subsection{Propose Evaluation Measures}
\subsubsection{Euclidean distance (ED)} We adopt the {\it Euclidean distance} (ED), which is an obvious choice of method to measure how far two points are in $n$-dimensional spaces. For this reason, we compute, for each instance of each data set and NN model, the ED between the point described in the GTE data and the point described through LIME's explanation coefficients. The range of the ED is (-$\infty$, +$\infty$), however, we do normalize the ED using the maximum and minimum points obtained for each data set and parameters. The goal is to keep the ED's results within the interval [0, 1] for better visualization. The purpose of computing the ED between the GTE data and LIME's explanation coefficients is to measure {\it accuracy} as a measure of {\it explanation goodness}. The ED, for being a distance, produces results in the opposite direction of quality. For this reason, later we will compute the {\it Complement of ED}, which we denote as {\it C-of-ED}, as its mathematical complement.

\subsubsection{Implementation Invariance}  \citet{sundararajan2017axiomatic} proposed that explanation methods should produce identical attributions for networks that produce the same outputs while receiving the same inputs, which are referred to as functionally equivalent. This is why we created models with different architectures for two of our data sets.

\subsubsection{Measures of Order} We propose to use the order of the explanation coefficients as measures of {\it accuracy} or {\it explanation goodness}. In LIME \citep{ribeiro2016should}, the explanation coefficients assign importance to each feature in the sense that the feature that is assigned the highest coefficient is the most important feature in the explanation. This is related to the use of gold features to evaluate explanations, as proposed by \citet{ribeiro2016should}. When gold features are used, the evaluation often targets the inclusion or not of a feature in an explanation. In the studies in this paper, we do not discuss the inclusion or not of features because our data sets have three, four, and five features each. At these small numbers of features, LIME includes all of them; this way we do not have to evaluate whether a feature is present, but how important it is considered. Note that the order of features is particularly proposed to evaluate LIME given the format LIME presents its explanations, although this would be an important aspect to consider when evaluating any explanation method. 

We define two evaluation measures of order: {\it Second Correct} and {\it All Correct}. Respectively, {\it Second Correct} indicates whether the second feature in the descending order of importance of an explanation's coefficients is correct in the sense that it is the same feature ordered as the second most important in the GTE data. Then {\it All Correct} indicates that all the features are in the same order as the features in the GTE data. The values for these measures are counted as 1 or 0 for each instance. The comparisons include results for 100 runs, hence the values represent percentages.
\subsection{Comparing GTEs to LIME Explanations}

\subsubsection{NN1 vs. NN2}
We start by comparing ED across the two different NN architectures for data sets Loan and Time, NN1 and NN2 to assess {\it Implementation Invariance}. Both were executed for 100 runs and thus we compute average and standard deviation across the 100 runs for each instance.

We use the parametric statistical testing t-test to measure whether the values differ significantly across the two samples NN1 and NN2. To conduct the t-test, we pose the hypothesis that the true difference between NN1 and NN2 is zero. The t-test determines that for $p$-values greater than 0.1, we cannot reject the hypothesis that the difference is zero between the samples. The $p$-values computed for Loan and Time data set are respectively, 0.979 and 0.661. These resulting $p$-values show that for both Loan and Time, the differences between NN1 and NN2 are not statistically significant. \citet{sundararajan2017axiomatic} suggests that explanation methods should satisfy {\it Implementation Invariance} for functionally equivalent NNs. This means that their explanations ought to be the same. As far as the t-test shows, the explanation coefficients are not significantly different, so at this level of specificity, they satisfy {\it Implementation Invariance}. Given these results, we will use only NN1 for the remainder of the studies.

\begin{figure}
\centerline{\includegraphics[width=\columnwidth, height=6cm]{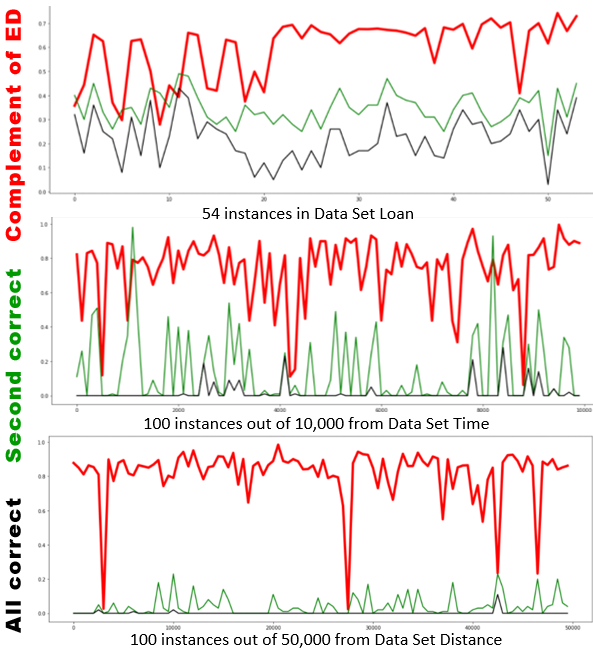}} 
\caption{C-of-ED (red), Second Correct (green), and All Correct (black) for data sets Loan, Time, and Distance}
\label{CofEDvs}
\end{figure}

\begin{table}[t]
\centering
\begin{tabular}{p{0.1\textwidth} |c|c|c c}
    \hline
   Data Set & Ave. C-of-ED & Ave. Second & Ave. All  \\
    \hline
  Loan &  0.47 & 0.32 &  \textbf{0.179}\\
    \hline
  Time &  0.76 & 0.03 &  0.0008\\
    \hline
  Distance & \textbf{0.82}  & \textbf{2.88} &  0.08\\      
\end{tabular}
\caption{Averages (Ave.) obtained for C-of-ED, {\it Second correct}, and {\it All correct} for all 100 runs and all instances for NN1 for the three data sets}
\label{5-25-50}
\end{table}

\subsubsection{Comparing C-of-ED against Second Correct and All Correct}
Now we compare the measures of order {\it All Correct} and {\it Second Correct} against the complement of the ED, {\it C-of-ED}. We use the complement given that measures of order are in the opposite direction of ED. Figure \ref{CofEDvs} shows the three measures for Loan, Time, and Distance. 

The visual inspection suggests multiple ideas. First, if we look at the red line for {\it C-of-ED}, it shows that the quality of LIME's explanations seem to increase with larger values of the number of samples hyperparameter. Note that the charts for Time and Distance show only 100 samples because showing all 10,000 and 50,000 would make them indecipherable. For this reason, we include Table \ref{5-25-50} with the averages for all instances to help the interpretation of the charts. The averages for Loan, Time, and Distance are, respectively, 0.47, 0.76, and 0.82. Recall the numbers set for number of samples submitted to LIME for these respective data sets were respectively, 25, 1,000, and 5,000. This seems reasonable as it allows LIME more chances to populate the region of the instance to be explained, thus increasing its chances of success.

Second, the measure {\it All Correct} (black line) in Figure \ref{CofEDvs} represents the number of times an instance has all its feature coefficients in the same order as GTE's coefficients. It is not surprising that this is the line further down with respect to the $y$-axis as it is the more demanding than {\it Second Correct} (in green). 

Third, even with limited samples in the charts for Time and Distance, we see that the quality of LIME's explanations vary. This deserves a more detailed analysis. With the small data set Loan, we can see, for example, all three measures agree that Instance 50 is lower in quality than Instance 49. But before we examine the numbers and explore potential reasons for LIME having more difficulty to explain some instances over others, let's scrutinize these measures.

\subsection{Validating Evaluation Measures}
In this section, we investigate whether we have any evidence to support these results by further analyzing what these measures above can tell us about LIME explanations. To do this, we now focus on the Data Set Loan because its small scale allows us to conduct a detailed and comprehensive analysis. Above when we described the experimental design for the Loan data, we mentioned we selected the parameter number of samples to be 25. We now expand the results for two more number of sample values, 5 and 50.

\begin{figure}[t]
\centerline{\includegraphics[width=\columnwidth, height=6cm]{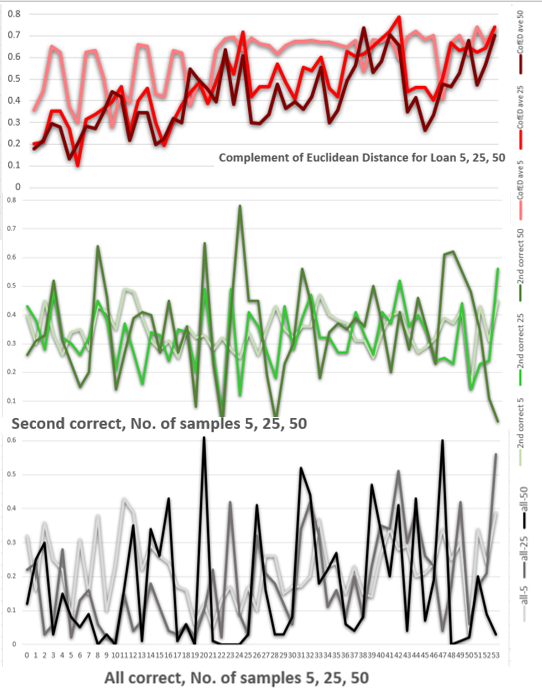}} 
\caption{Comparison of C-of-ED (top), Second Correct (middle), and All Correct (bottom) for Data Set Loan and number of samples 5, 25, and 50} 
\label{ed-5-25-50}
\end{figure}

Figure \ref{ed-5-25-50} shows the measures {\it C-of-ED}, {\it Second correct}, and {\it All correct} for the different hyperparameter number of samples used with the Loan data. We kept the colors we used in earlier charts, making lighter hues for number of samples 5, darker for 50, and kept an intermediary tone for 25. For C-of-ED, the average at 5 number of samples is the highest, 0.60 against 0.47 and 0.41 for 25, and 50. For {\it Second correct}, the highest average is 0.35, obtained with 5 and 50 number of samples, against 0.32 with 25. For {\it All correct}, the highest is again at 5 number of samples with 0.22 against 0.18 and 0.16 for 25 and 50.  

The first observation is that these results do not match the conclusion above that higher number of samples lead to more accurate explanations although this observation makes sense technically. Thorough examination of the results for every instance reveals that, at 5 number of samples, data representing GTE has a very high proportion of coefficients that are zero. The exact number is 18 zeros for coefficient $x_1$, 23 for $x_2$, and 27 for $x_3$, corresponding to 18, 23, and 50\%. These high numbers of zeros can be explained by the low number of samples that would make it hard to fit the linear regression and thus return zeros. We then examined the number of zeros in the coefficients produced by LIME and we noted that in the 100 runs of 54 instances, the total number of zeros is 26, 27, and 29, respectively for $x_1$, $x_2$, and $x_3$, representing averages of 2.6, 2.7, and 2.9 for all 54 instances (these are around 5\%). Consequently, given that LIME coefficients do not have such an abundance of zeros, ED will artificially show better results at number of samples 5 because these distances will be shorter. A distance between a number, which can be positive or negative, and zero will be shorter than a number and another that may also be positive or negative. The zeros also cause problems in computing the measures of order.

Two observations can be made from the identification of these high volumes of zero. First, the evaluations for 5 number of samples for the Loan data are artificial. They are revealed by the measures as good but the numbers are artificial, they do not originate from better explanations. Consequently, we do not have any reason to question that higher number of samples lead to better quality explanations.

Second, these artificially produced number do indicate better quality and all the proposed measures have shown them. This supports the quality of the proposed measures.

Finally, these studies suggest that the best quality of explanations from LIME for the Loan Data should be when using 50 number of samples, but the measures do not show this with consistency. Consider that 50 number of samples is almost as much as the total number of instances in the Data Set Loan. With both the data representing GTEs and LIME using 50 number of samples, what would be the cause of the difference in the coefficients? If we could tell LIME the range and precision of the allowable values for the data to use in the perturbations, with only three features and a NN with 100\% accuracy, LIME would only generate perturbations that matched the actual data set, and given that we used the same cosine similarity and the same Ridge regression with the same parameters, LIME's perturbations would be all actual instances. When using 50, it would be 50 out of 54, exactly like the data representing GTEs. Consequently, the only point of information separating LIME from better ({\it i.e}., more accurate) explanations is not knowing the range and precision of allowable values. In practice, in a real-world model that needs explanation, there is nothing preventing us from asking the actual values allowed in data to create more accurate perturbations. This demonstrates how the use of data representing ground-truth explanations can lead to analyses that will improve existing XAI methods. 

\section{Discussion and Conclusions}
The methodology we describe to generate data representing {\it ground-truth explanations} (GTEs) poses many challenges. It requires the identification of a data-generation process and needs equations to define classes. The possibility of class overlap, their benefits and limitations, and methods to avoid noise are questions for future work.

The need to align data representing GTEs with the method targeted to evaluate may pose challenges such as the one we faced when setting a low value to a hyperparameter that produced artificially good results. This suggests this approach may be far from being fully automated.

The proposing authors of {\it implementation invariance} \cite{sundararajan2017axiomatic} suggest that explanation methods should satisfy it, which means producing the same explanation as long as NNs are functionally equivalent. If we envisage an explanation in support of accountability reports, then we want to have methods that can distinguish when a different architecture leads to a different explanation. Furthermore, when computing {\it implementation invariance}, we face the question of which level of specificity to compare the explanations from these models. This poses questions with respect to whether what being the same for explanations mean. Consider that this question will differ depending on how the XAI method formats explanations.

We analyzed the results of our evaluation of LIME and showed how that analysis led us to conclusions about how LIME could be improved. Although not explicitly shown, our proposed method is measurable and verifiable, allowing the comparison between two explanation approaches. Further work examining why a method performs better in a certain type of instance, such as in outliers vs. non-outlier instances can help direct how to improve said methods. Finally, this proposed approach sheds light into how to demonstrate accountability, to create benchmarks, and contribute to the progress of the field.

All data and code necessary for reproducibility is available at https://github.com/Rosinaweber forwardslash DataRepresentingGroundTruthExplanations/tree/master.

\subsubsection{Acknowledgements}
Rosina Weber and Prateek Goel are supported by the National Center for Advancing Translational Sciences (NCATS), National Institutes of Health, through the Biomedical Data Translator program award {\#}OT2TR003448. Any opinions expressed in this document are those of the authors and do not necessarily reflect the views of NCATS, other Translator team members, or affiliated organizations and institutions.

\bibliography{amiri-aaai.bib}
\end{document}